%% file: example.tex
\documentclass{article}
\usepackage{bigfoot}
\usepackage{float}
\usepackage{booktabs}
\usepackage{amsthm}
\usepackage{amsfonts}
\usepackage{graphicx}
\usepackage{amsmath}
\usepackage{tikz}
\usetikzlibrary{arrows.meta, positioning, fit, backgrounds, calc}
\usepackage[final]{corl_2026} % Uncomment for the camera-ready ``final'' version.
\usepackage{cleveref}
\usepackage{enumitem}
\usepackage{subcaption}
\newtheorem{proposition}{Proposition}

\title{TARC: Time-Adaptive Robotic Control}

\author{
 Arnav Sukhija\thanks{Correspondence to: \texttt{asukhija@ethz.ch}}, \; Lenart Treven, \; Jin Cheng, \;
 Florian Dörfler, \; Stelian Coros, \; Andreas Krause \\[6pt]
   \normalsize
  ETH Zurich \\
  \url{https://arnavsukhija.github.io/projects/tarc}
}

\begin{document}
\maketitle

%===============================================================================

\begin{abstract}
\looseness=-1
Most robotic systems rely on fixed-frequency discrete-time controllers, creating
a trade-off between the efficiency of low-frequency control and the responsiveness of high-frequency feedback. As a result, systems typically default to high control rates for robustness, at the cost of wasted inference and unnecessary actuation. Addressing this, we introduce Time-Adaptive Robotic Control (TARC), a reinforcement learning framework in which the policy jointly predicts a control action and its duration of application. TARC learns temporally extended actions by optimizing task performance under soft or hard constraints on the number of control switches, enabling adaptive modulation of control rates.
\looseness=-1
We evaluate TARC on two robotic hardware platforms: a high-speed RC car and the Unitree Go1 quadruped, and on a vision-language action model in simulation, where each query incurs a costly transformer forward pass. Across all settings, TARC matches the performance of high-frequency discrete-time controllers while operating at less than half their control frequency. Unlike fixed-rate controllers, TARC adapts its control frequency online, allocating high-frequency feedback only when required.
\end{abstract}


%===============================================================================
\input{introduction}

%===============================================================================
\input{relatedwork}

%===============================================================================

\input{method}
%===============================================================================

\input{experimentalSetup}
%==============================================================================

\input{results}

%===============================================================================

\input{conclusion}
 
%===============================================================================

\clearpage
% The acknowledgments are automatically included only in the final and preprint versions of the paper.
\acknowledgments{
This project has received funding from the Swiss National Science Foundation under NCCR Automation, grant agreement 51NF40 180545.}

%===============================================================================

\bibliography{example}  % .bib

% -------------------------------------------------------
\clearpage
\appendix

% ----------------------------------------------------------

\input{appendixA}
% -------------------------------------------------------
\input{appendixB}
% =====================================================================
\input{appendixC}

 % ---------------------------------------------------------------------

\end{document}

%% file: introduction.tex
\section{Introduction}
%\vspace{-3pt}

Most robotic systems rely on fixed-frequency discrete-time controllers, creating a fundamental trade-off between the efficiency of low-frequency control and the responsiveness of high-frequency control. In practice, systems default to high control rates to handle worst-case demands: a quadruped runs its locomotion policy at the same rate when standing still as when recovering from a push; a self-driving car queries its planner at the same rate when cruising as when executing an evasive maneuver. This conservative choice is robust but wasteful: it incurs unnecessary inference cost and actuation effort during easy states~\cite{10342408, wang2024moseacstreamlinedvariabletime}. The cost
is especially pronounced for modern vision-language-action (VLA) models, where each policy query involves a costly transformer forward pass, making control frequency a critical deployment constraint.

%A natural alternative is to let the controller modulate its own frequency in
%response to situational demand. Adaptive frequency control has received attention in recent learning-based 
%work~\cite{wang2024moseacstreamlinedvariabletime, 
%wang2024reinforcementlearningelastictime, treven2024sensecontroltimeadaptiveapproach} 
%and in model-predictive control~\cite{10806807, li2025hold}, but existing approaches have not demonstrated principled, budget-aware frequency adaptation on dynamic real-world robotic platforms.

A natural alternative is to let the controller modulate its own frequency in response to situational demand. Prior work has explored this direction by incorporating frequency into the reward function or learning to repeat actions over variable durations~\cite{wang2024moseacstreamlinedvariabletime,
wang2024reinforcementlearningelastictime,
park2022timediscretizationinvariantsafeaction, pmlr-v119-metelli20a,
treven2024sensecontroltimeadaptiveapproach}. While these methods demonstrate that adaptive frequency is beneficial, they require task-specific reward redesign to encode frequency preferences and have not demonstrated budget-aware adaptation on dynamic real-world platforms.

%\vspace{-2pt}
In this work, we present \textbf{Time-Adaptive Robotic Control (TARC)}, a reinforcement
learning approach in which policies jointly select a control action and the duration for which it is applied. We formulate adaptive frequency control as a constrained Markov Decision Process (MDP): maximize task reward subject to a budget on the expected number of control switches. %The Lagrangian relaxation yields a switch cost that can be treated as a soft or hard constraint. Specifically, we contribute:
Our key contributions are as follows.
\begin{itemize}[topsep=2pt,itemsep=5pt,parsep=0pt,leftmargin=1.4em]
    \item \textbf{A constrained MDP formulation} of adaptive frequency control with a principled Lagrangian relaxation that can be combined with any standard RL algorithm to learn time-adaptive controllers. %that unifies soft and hard on the frequency of querying the policy. %Under the hard constraint,
    %dual gradient ascent tracks user-specified switch budgets with mean
    %deviation below $5\%$, removing the need for per-platform reward engineering.
    \item \textbf{Zero-shot sim-to-real deployment} on two dynamically
    distinct hardware platforms: a high-speed RC car and a Unitree Go1
    quadruped. TARC matches or exceeds high-frequency baselines in task
    reward while operating at less than half their control frequency, reducing onboard policy inference by $64\%$ and motor command travel by $21\%$.
    %\item %\textbf{Evidence that adaptivity itself, not merely a lower average rate,
    %drives these gains.} A fixed controller trained at TARC's \emph{average} frequency
    %performs comparably, but a fixed controller at TARC's \emph{minimum} frequency does
    %not: TARC substantially outperforms it by spending its switch budget where the task
    %demands reactivity.
    \item \textbf{Time-adaptive control for VLA policies}, where TARC reduces the transformer forward passes by approximately $38\%$, while maintaining the task performance of the base VLA.
\end{itemize}

\begin{figure}[t]
    \centering
    \includegraphics[width=0.9\columnwidth]{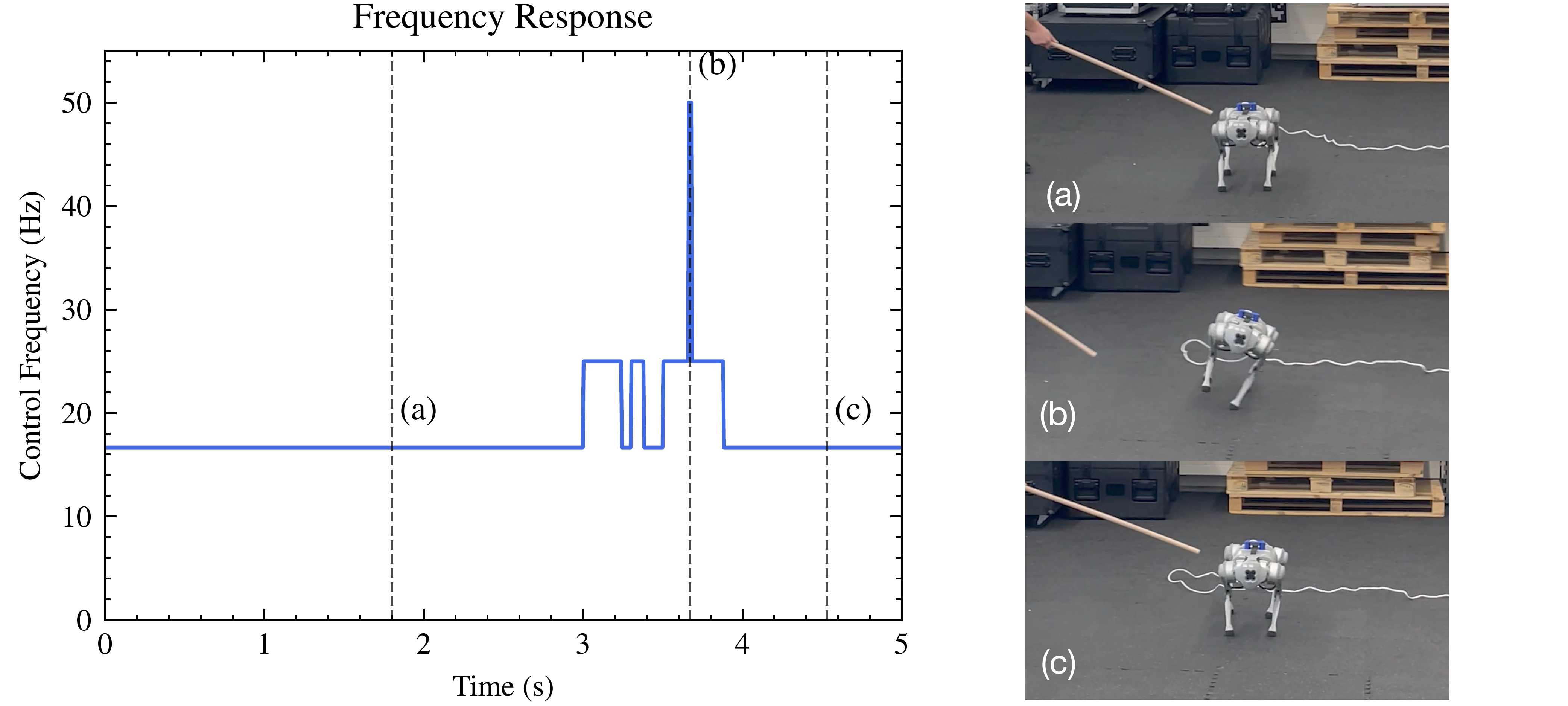}
   % %\vspace{-6pt}
    \caption{Control frequency of TARC during a perturbation experiment on the
    Go1. During stable standstill (a) the policy operates at $16.7$\,Hz; upon
    being pushed (b) it spikes up to $50$\,Hz to handle the disturbance, then
    recovers to low frequency once the robot stabilises (c).}
    \label{fig:perturbations}
\end{figure}

Figure~\ref{fig:perturbations} illustrates this adaptive modulation behavior on the Unitree Go1. 

%% file: relatedwork.tex
\section{Related Work} 
\looseness=-1 
\textbf{Choice of Control Frequency in Robotic Systems} Traditional robotic control schemes often operate at a fixed frequency, which introduces several practical limitations. Running at the worst-case rate incurs unnecessary inference cost and actuation effort during easy states, a concern explicitly raised in the context of real-robot deployment by~\citet{wang2024reinforcementlearningelastictime}. Fixed-frequency controllers can also be wasteful if the frequency is too high, and their training limitations have been extensively studied: \citet{park2022timediscretizationinvariantsafeaction} and \citet{pmlr-v139-amin21a} show that variable action durations can significantly improve learning performance. Although high frequencies ensure reactivity, they incur significant computational costs and can hinder training~\citep{10342408, wang2024moseacstreamlinedvariabletime}. In contrast, low-frequency control is efficient and can improve latency insensitivity~\citep{10160357}, but risks instability when rapid reactions are needed. \citet{10806807} challenge the intuition of ``one frequency for all'' by using different control frequencies for different phases of a dynamic jumping maneuver on legged robots, observing more robust results with phase-specific rates. Similarly, \citet{li2025gaitnetaugmentedimplicitkinodynamicmpc} use a two-frequency agent framework for humanoid locomotion, collectively motivating adaptive control strategies. 

\textbf{Adaptive Query Frequency Control} \looseness=-1 To overcome these limitations, recent research has explored methods that adapt the control frequency at run-time. A significant body of work on repetitive action reinforcement learning enables agents to repeat actions over multiple time steps~\citep{Hafner2020Dream, braylan:aaai15ws, sharma2017learning, pmlr-v119-metelli20a}. Most closely related are MOSEAC~\citep{wang2024moseacstreamlinedvariabletime}, which extends SAC with variable time steps, and elastic time steps~\citep{wang2024reinforcementlearningelastictime}, which combine multi-objective reward shaping with variable duration selection. Both jointly learn actions and execution durations, but encode the timing preference in a reward that must be re-derived per platform, and have been validated primarily on navigation tasks. These methods price interaction through a cost, leaving the realized rate an outcome of weight tuning. TARC instead formulates query-rate control as a Constrained Markov Decision Process (CMDP), which puts the constraint on the rate itself: the Lagrangian relaxation adds only a single scalar to the existing reward. Furthermore, we demonstrate zero-shot sim-to-real transfer on a drifting RC car and a quadrupedal robot.  

A related paradigm in robot learning is action chunking~\citep{zhao2023learningfinegrainedbimanualmanipulation}, where policies output fixed-length open-loop action sequences, widely adopted in diffusion policies~\citep{chi2024diffusionpolicyvisuomotorpolicy} and vision-language action  models~\citep{black2026pi0visionlanguageactionflowmodel}. A fixed chunk size trades query frequency against responsiveness~\citep{liang2026adaptiveactionchunkinginferencetime}; recent work adapts chunk size at inference time using action entropy~\citep{liang2026adaptiveactionchunkinginferencetime}. In contrast, TARC operates at the post-training stage of the VLA. Specifically, we leverage diffusion steering~\citep{wagenmaker2025steeringdiffusionpolicylatent} to guide policy generation while additionally predicting how much of an action chunk should be executed before re-querying the VLA. This allows the system to selectively increase control frequency when rapid reactivity is required by the task.

The question of \emph{when} to issue a control update also connects to
event- and self-triggered control~\citep{6425820, 5411835}, where
updates are executed only when predefined stability conditions are
violated. These methods minimize unnecessary interventions but rely on
hand-crafted triggering conditions; TARC instead acquires its update
policy entirely from task reward.

%% file: method.tex
\section{Method} \label{Methodology}
%%\vspace{-3pt}

We seek a control policy that adapts its control frequency to situational demand, operating at high frequency when the task requires rapid corrections and at low frequency otherwise. We formulate this as a constrained optimization problem and derive a practical RL objective from its Lagrangian relaxation. 

% Drop this figure into the Method section, e.g. after the Policy class paragraph.

\begin{figure}[t]
    \centering
    \resizebox{0.82\textwidth}{!}{%
    \begin{tikzpicture}[
        font=\sffamily,
        phase/.style={draw, rectangle, dashed, inner sep=0.45cm, rounded corners=3pt},
        arrow/.style={-Latex, thick}
    ]
        % ---- Axes ----
        \draw[->, thick] (0,0) -- (9.2,0)
            node[right, font=\footnotesize\sffamily] {time};
        \node[rotate=90, font=\footnotesize\sffamily, anchor=south]
            at (-0.35,0.45) {$u_t$};

        % ---- High-frequency region (Delta t = 1) ----
        \draw[red!70!black, line width=3pt] (0.0,0.45) -- (0.8,0.45);
        \draw[red!70!black, line width=3pt] (0.8,0.72) -- (1.6,0.72);
        \draw[red!70!black, line width=3pt] (1.6,0.52) -- (2.4,0.52);

        % ---- Medium-frequency region (Delta t = 2) ----
        \draw[orange!80!black, line width=3pt] (2.4,0.62) -- (4.0,0.62);
        \draw[orange!80!black, line width=3pt] (4.0,0.48) -- (5.6,0.48);

        % ---- Low-frequency region (Delta t = 4) ----
        \draw[blue!70!black, line width=3pt] (5.6,0.58) -- (8.8,0.58);

        % ---- Vertical dividers at query events ----
        \foreach \x in {0.8,1.6,2.4,4.0,5.6,8.8}{
            \draw[dashed, gray!60, thin] (\x,-0.08) -- (\x,1.0);
        }

        % ---- Query dots on time axis ----
        \foreach \x in {0.0,0.8,1.6,2.4,4.0,5.6}{
            \filldraw[gray!60] (\x,0) circle (2.2pt)
                node[below, font=\tiny\sffamily, yshift=-2pt] {query};
        }

        % ---- Duration brackets ----
        \draw[<->, thin] (0.0,-0.55) -- (0.8,-0.55)
            node[midway, below, font=\small\sffamily] {$\Delta t{=}1$};
        \draw[<->, thin] (2.4,-0.55) -- (4.0,-0.55)
            node[midway, below, font=\small\sffamily] {$\Delta t{=}2$};
        \draw[<->, thin] (5.6,-0.55) -- (8.8,-0.55)
            node[midway, below, font=\small\sffamily] {$\Delta t{=}4$};

        % ---- Region labels ----
        \node[font=\footnotesize\sffamily, red!70!black]
            at (1.2, 1.22) {high $f$};
        \node[font=\footnotesize\sffamily, orange!80!black]
            at (4.0, 1.22) {mid $f$};
        \node[font=\footnotesize\sffamily, blue!70!black]
            at (7.2, 1.22) {low $f$};

        % ---- Invisible anchors for outer box ----
        \node[inner sep=0pt] (BL) at (-0.7,-0.85) {};
        \node[inner sep=0pt] (TR) at ( 9.5, 1.55) {};

        \begin{pgfonlayer}{background}
            \node[phase, fit=(BL)(TR)] {};
        \end{pgfonlayer}

    \end{tikzpicture}%
    }
    \caption{TARC's adaptive query schedule. The policy is queried (gray dots) only
    when a new action is needed; between queries, the previous action $u_t$ is held for
    $\Delta t$ steps. During demanding phases the policy issues frequent queries (short
    $\Delta t$, \textcolor{red!70!black}{red}); during stable phases it holds actions
    longer (large $\Delta t$, \textcolor{blue!70!black}{blue}), reducing inference cost
    proportionally.}
    \label{fig:overview}
\end{figure}

%\vspace{-2pt}
\subsection{Problem statement}

Consider a discrete-time MDP with state space $\mathcal{X}$, action space
$\mathcal{U}$, transition dynamics $p(x_{t+1} \mid x_t, u_t)$, reward function
$r: \mathcal{X} \times \mathcal{U} \to \mathbb{R}$, and discount factor
$\gamma \in (0,1]$. A standard fixed-frequency policy
$\pi: \mathcal{X} \to \mathcal{U}$ issues a new action at every control step,
regardless of whether the situation demands it. We instead seek a policy that issues queries only when necessary,
subject to a budget on how often the policy is queried. Let $q_t \in \{0, 1\}$
indicate whether the policy is queried at step $t$. We formulate adaptive
frequency control as a constrained MDP:
\begin{equation} \label{eq:cmdp}
    \max_{\pi} \;\;
    \mathbb{E}_{\pi}\!\left[\sum_{t=0} \gamma^t r(x_t, u_t)\right]
    \quad \text{subject to} \quad
    \mathbb{E}_{\pi}\!\left[\sum_{t=0} \gamma^t q_t\right]
    \leq \frac{K}{1-\gamma},
\end{equation}
where $K \in (0,1]$ is a user-specified query budget. A fixed-frequency controller has $q_t = 1$ for all $t$, saturating the budget at $K=1$; a budget $K < 1$ requires the
policy to skip queries at some steps by holding previous actions. We next introduce the time-adaptive policy class as the mechanism that implements this.

\subsection{Time-Adaptive Policies}
\label{def:timeAdaptive}
\paragraph{Policy Class.}
A \emph{time-adaptive policy} $\pi$ maps the augmented state $s_t = (x_t, t)$ to a joint action-duration pair:
\begin{equation}
    \pi(s_t) = (u_t, \Delta t), \quad
    u_t \in \mathcal{U}, \Delta t \in \{1, \ldots i\}
\end{equation}
where $\Delta t$ is the number of steps for which $u_t$ is held constant,
and $i \in \mathbb{N}$ is a hyperparameter controlling the maximum repetition
count. An action held for $\Delta t$ steps contributes at most one query
over $\Delta t$ time steps. The effective control frequency varies as
\begin{equation}
    f = \frac{f_{\max}}{\Delta t} \in \left[\frac{f_{\max}}{i},\, f_{\max}\right],
\end{equation}
with larger $i$ allowing lower minimum frequencies at the cost of longer
open-loop intervals. \Cref{fig:overview} provides an overview of this formulation.

\paragraph{Lagrangian relaxation.}
Introducing a multiplier $\lambda \geq 0$ for the constraint in
\cref{eq:cmdp} yields the penalized objective
\begin{equation} \label{eq:lagrangian}
    \mathcal{L}(\pi, \lambda) =
    \mathbb{E}_{\pi}\!\left[\sum_t \gamma^t r(x_t, u_t)\right]
    - \lambda \!\left(\mathbb{E}_{\pi}\!\left[\sum_t \gamma^t q_t\right]
    - \frac{K}{1-\gamma}\right).
\end{equation}

\begin{proposition}[Per-step decomposition]
\label{prop:decomp}
For a time-adaptive policy $\pi$, the Lagrangian $\mathcal{L}(\pi, \lambda)$ decomposes into a sum of
per-decision rewards. Specifically, optimizing $\mathcal{L}$ is equivalent
to optimizing the augmented MDP with per-decision reward
\begin{equation} \label{eq:reward}
    R(s_t, a_t) =
    \left(\sum_{k=0}^{\Delta t - 1} \gamma^k r(x_{t+k}, u_t)\right) - c,
\end{equation}
where $a_t = (u_t, \Delta t)$ is the augmented action and $c = \lambda$.
\end{proposition}
%\vspace{-2pt}
\noindent\textit{Proof sketch.}
Each policy query at decision time $t$ contributes $q_t = 1$ to the query count and
accumulates reward over the held interval. Substituting the time-adaptive policy class
into \cref{eq:lagrangian} and grouping terms by decision epochs yields a telescoping sum in which each epoch contributes exactly \cref{eq:reward} plus a constant independent of $\pi$. The full derivation is given in Appendix~\ref{app:proof}.

%\vspace{-2pt}
\subsection{Soft and Hard Query Budget Constraints} \label{sec:constraints}

The multiplier $\lambda$ (equivalently $c$) can be treated in two ways, yielding qualitatively different operating modes. 

\paragraph{Soft constraint (fixed $c$).}
Fixing $c$ and optimizing \Cref{prop:decomp} directly encourages sparser control without enforcing a precise query rate $K$. The cost acts as a directional signal: larger $c$ produces lower average frequency, but the realized number of queries is not directly constrained. This variant is similar to \citet{treven2024sensecontroltimeadaptiveapproach}, who price interactions through a soft cost in continuous time. We differ in two ways: we
study the discrete-time RL setting that robot control stacks and VLA chunking operate,
and the hard-constraint variant below targets a query rate directly rather than a cost. We study how a soft-constrained cost $c$ affects the performance and control frequency of the RL policy in Appendix~\ref{app:switchcost}. 

\paragraph{Hard constraint (auto-tuned $\lambda$).}
To enforce a precise budget $K$, we treat $\lambda$ as a learnable parameter updated by dual gradient ascent alongside policy optimization: 
\begin{equation} \label{eq:dual}
    \lambda_{k+1} = \max\!\left(0,\;
        \lambda_k + \alpha_\lambda\!\left(
        \mathbb{E}_{\pi_k}\!\left[\sum_t \gamma^t 
        q_t\right] - \frac{K}{1-\gamma}
        \right)\right),
\end{equation}
where the expectation is estimated over training batch and $\alpha_\lambda$ 
is the dual step size. When the discounted query count exceeds the budget, 
$\lambda$ increases, strengthening the penalty; when the policy is conservative, $\lambda$ decreases. This closes the loop between 
the theoretical constraint in \cref{eq:cmdp} and the practical training 
objective in \cref{eq:reward}. We show in \Cref{sec:sim} that this 
procedure reliably tracks user-specified budgets $K$ across a range of values for $K$.

%% file: experimentalSetup.tex
\section{Experimental Setup}
\label{sec:setup}
%\vspace{-2pt}

We evaluate \textbf{TARC} across three platforms. On the RC car and Unitree Go1, we
compare against fixed-frequency baselines at $f_{\max}$ and deploy zero-shot on
hardware; on the VLA, evaluation is in simulation where each policy query incurs a full
transformer forward pass. The maximum hold duration $i$ and query cost $c$ are set per
platform; sensitivity to both is in Appendix~\ref{app:switchcost}. The RC car and Go1 policies
are optimized with
PPO~\cite{schulman2017proximalpolicyoptimizationalgorithms,
freeman2021braxdifferentiablephysics}. Results report mean and standard error over 5
seeds unless noted otherwise. Full hyperparameters are in Appendix~\ref{app:hyperparams}.

Throughout, \emph{unpenalized reward} denotes the task reward with the query cost
zeroed at evaluation, and \emph{penalized reward} subtracts $c \times (\text{query
count})$, using the same $c$ for every method, fixed-frequency baselines included.

Our experiments address three questions: \textbf{(i)} Can TARC reliably track a user-specified query budget, and how does task performance vary with the budget level? \textbf{(ii)} Does operating at lower average control frequency come at a cost in task reward, and does adaptive frequency outperform a fixed controller at the same average rate? \textbf{(iii)} Does the approach generalize beyond locomotion to large generative policies such as VLAs, where each query incurs a costly transformer forward pass?

\begin{figure}[ht]
    \centering
    % Left: RC Car
    \begin{subfigure}{0.32\linewidth}
        \centering
        \includegraphics[width=\linewidth]{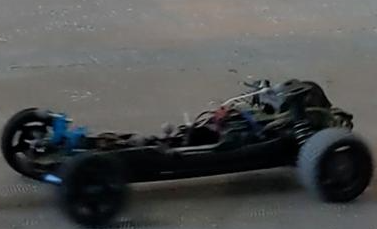}
        \caption{Radio-Controlled (RC) car.}
        \label{fig:setup_car}
    \end{subfigure}
    \hfill
    \hfill
    % Middle: Unitree Go1
    \begin{subfigure}{0.32\linewidth}
        \centering
        \includegraphics[width=\linewidth]{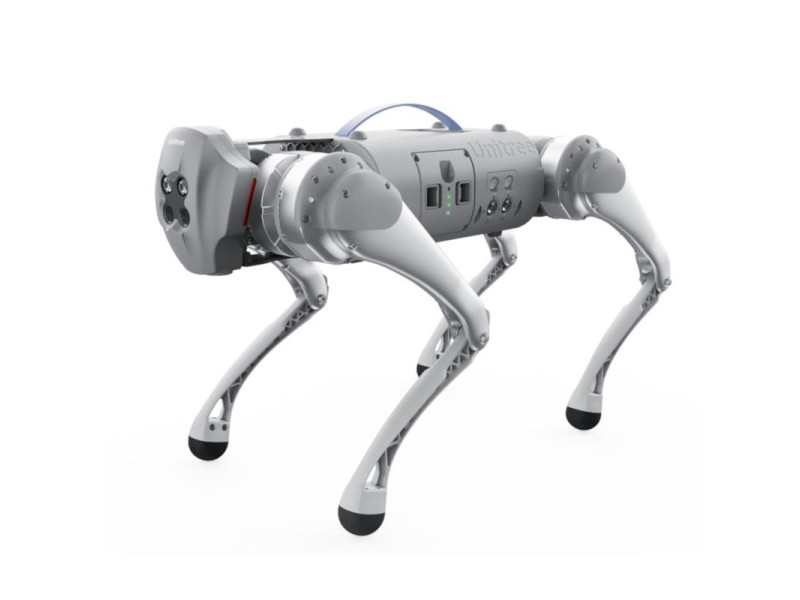}
        \caption{Unitree Go1 quadruped.}
        \label{fig:setup_quadruped}
    \end{subfigure}
        % Right: VLA
    \begin{subfigure}{0.32\linewidth}
        \centering
        \includegraphics[width=0.75\linewidth]{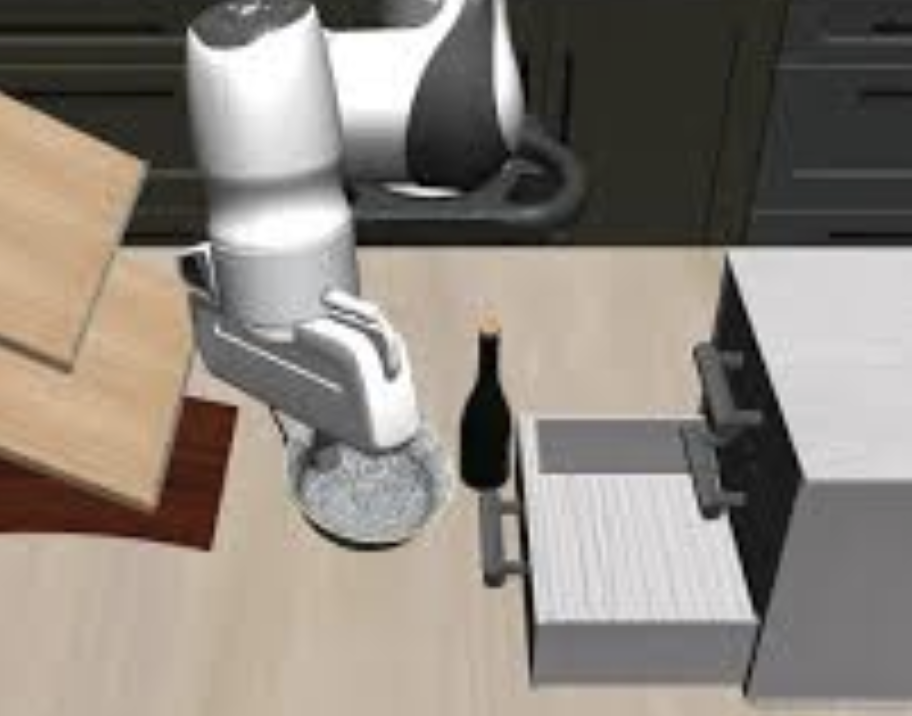} % Remember to crop the top wall!
        \caption{LIBERO Benchmark with pick-and-place tasks.}
        \label{fig:setup_arm}
    \end{subfigure}
    
    \caption{Platforms used in the experimental evaluation: (a) RC car, (b) a quadrupedal robot, and (c) LIBERO benchmark.}
    \label{fig:experimental_setup}
\end{figure}
 
\textbf{RC Car: } We use an RC car~\cite{bhardwaj2024dataefficienttaskgeneralizationprobabilistic,
sukhija2023gradientbasedtrajectoryoptimizationlearned} with $f_\text{max} = 30$\,Hz and for state estimation, we use the OptiTrack Motion Capture System~\cite{OptiTrackRobotics}. The simulator uses the Pacejka tire model~\cite{Pacejka01011992, kabzan2019amzdriverlessautonomousracing} with domain randomization over mass, tire stiffness, and motor
parameters~\cite{tobin2017domainrandomizationtransferringdeep} (see Appendix~\ref{app:rccar}). The task is a reverse-parking maneuver: starting roughly 2m from the target, the car must rotate $180^\circ$ and park, which usually requires a drift. Episodes run for 200 steps ($\approx$6.6\,s). The reward follows~\citet{rothfuss2024bridgingsimtorealgapbayesian}:
\begin{equation}
    r(x_t, a_t) = r_\text{state}(x_t) - w \cdot \|a_t\|, \quad w = 0.005,
    \label{eq:reward_rccar}
\end{equation}
where $r_\text{state}$~\cite{tassa2018deepmindcontrolsuite} peaks at the target pose and decays smoothly with distance. %We use $i=4$ and $c=0.1$.

\textbf{Unitree Go1: } We use the Unitree Go1~\cite{UnitreeGo1} (48-D state, 12-D action) with
$f_{\max} = 50$\,Hz, trained and deployed via MuJoCo
Playground~\cite{zakka2025mujocoplayground} on the JoystickFlatTerrain
environment. The reward function, domain randomization, and the low-level PD
controller are kept identical across all agents, so performance differences
are attributable solely to the time-adaptive action interface. TARC solely augments the observation with the episode timestep $t$, the network architecture and hyperparameters are kept identical between all agents. 
We test the policy on three 20-second scenarios not seen during training (Appendix~\ref{app:go1_scenarios}): \textbf{Gentle Curve} (smooth low-speed turning), \textbf{Velocity Changes} (straight walking at varying speeds), and \textbf{Run Then Turn} (running followed by an abrupt 
switch to in-place turning).

\textbf{VLA Model: }We evaluate TARC on the LIBERO benchmark~\cite{liu2023liberobenchmarkingknowledgetransfer}, specifically
the \texttt{LIVING\_ROOM\_SCENE3} pick-and-place task from the
\texttt{libero\_90} suite ($256{\times}256$ RGB
observations, sparse binary success signal, 400-step episode
horizon). As the base controller we use the publicly released $\pi_0$ checkpoint fine-tuned on LIBERO~\cite{black2026pi0visionlanguageactionflowmodel},
which emits action chunks of up to length 50; this sets the upper bound on TARC's adaptive replanning interval. We build on the DSRL
framework~\cite{wagenmaker2025steeringdiffusionpolicylatent} for $\pi_0$ fine-tuning, training a lightweight
MLP diffusion steering on top of the frozen $\pi_0$ backbone. 
%that adapts when to replan while leaving action generation unchanged. 
As baseline we use the fixed query frequency $\text{QF}{=}20$ as proposed by \citet{wagenmaker2025steeringdiffusionpolicylatent}. This corresponds to replanning every 20 steps. For TARC, we predict the frequency of replanning. 
We report the success rate over 10 rollouts per checkpoint.

%% file: results.tex
\section{Results} \label{results}
%\vspace{-2pt}
 
We structure our results into simulation studies (\Cref{sec:sim}) and
zero-shot hardware deployment (\Cref{sec:hardware}). In simulation we
analyze query budget enforcement, the benefit of adaptive over fixed
low-frequency control, and the extension to a vision-language action
model; on hardware we deploy our trained policies zero-shot on the RC
car and Unitree Go1. Additional ablations, e.g., sensitivity to query
cost $c$ and repetition bound $i$, are provided in Appendix~\ref{app:supplementary}. 

We also report significance tests with the training seed as the analysis unit. Unless stated otherwise, all methods use 5 seeds on the same fixed episode set, so the seed is the only random unit: we bootstrap over seeds and cross-check with an exact permutation test on the per-seed means, for which $p = 0.008$ is the smallest attainable value; on hardware, permutations are performed within scenario. Where we report that TARC matches a baseline in task reward, we test equivalence with two one-sided tests (TOST) and a margin fixed in advance at 5\% of the baseline mean, since a large p-value for a difference does not establish equivalence. Throughout, $\Delta$ denotes TARC's reward minus the baseline's.

%===============
%\vspace{-2pt}
%\subsection{Automatic Query Cost Selection}
%\label{sec:autotuning}
\subsection{Simulation Results}
\label{sec:sim}

\paragraph{Automatic query budget enforcement.}
\begin{figure}[ht]
    \centering
    \includegraphics[width=0.85\textwidth]{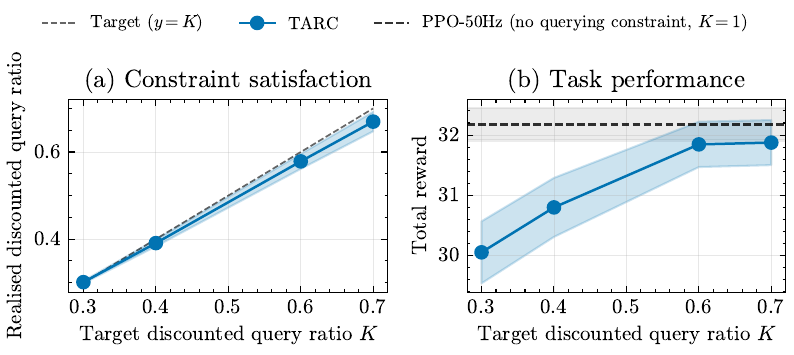}
    \caption{Automatic query budget enforcement on the Go1.
    \textbf{(a)}~Realized discounted query ratio tracks the target $K$
    with mean deviation below $5\%$ across all targets. \textbf{(b)}~Task
    reward varies smoothly and predictably with $K$.}
    \label{fig:autotuning}
\end{figure}
The hard-constraint variant enforces a user-specified query budget $K$
via dual gradient ascent (\Cref{eq:dual}). We train TARC on the Go1
Joystick environment across budgets $K \in \{0.3, 0.4, 0.6, 0.7\}$ and
measure the realized discounted query count against the target.
\Cref{fig:autotuning} shows that the realized query rate tracks $K$
closely across all budget levels. Task reward varies smoothly with $K$:
at tighter budgets ($K = 0.3$--$0.4$), reward falls below the
unconstrained PPO-50\,Hz baseline, since the policy is prevented from
querying as often as the task would otherwise demand. At $K \geq 0.6$,
reward recovers to within the baseline's confidence interval. This yields
a predictable and controllable performance-frequency trade-off.

\paragraph{Adaptive vs. fixed low-frequency control.}
\begin{figure}[ht]
    \centering
    \includegraphics[width=0.9\textwidth]{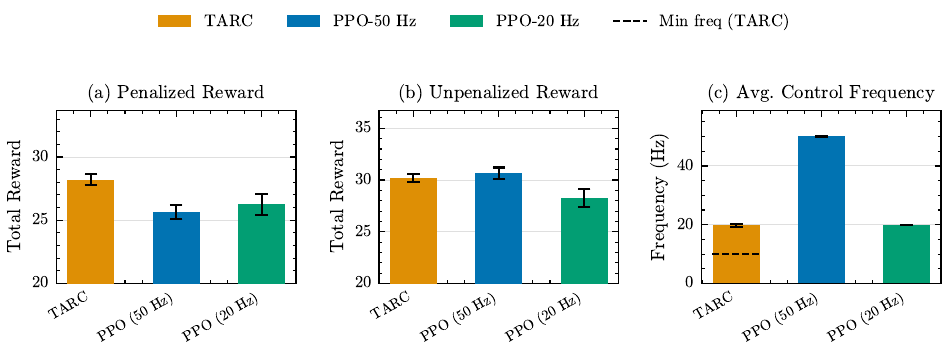}
    \caption{Adaptive (TARC) vs. fixed-frequency control on the Go1 in
    simulation. Panels (a)--(b) show penalized and unpenalized reward
    (y-axis truncated at 20 to highlight differences); (c) shows average
    control frequency. TARC matches the high-frequency PPO-50\,Hz baseline in unpenalized reward and attains comparable mean reward to the PPO-20\,Hz baseline trained at TARC's average operating frequency with lower variance.}
    \label{fig:go1_lowFreq}
\end{figure}
A natural question is whether TARC's benefit comes from adaptive
allocation or merely from operating at a lower average rate. To isolate
this, we compare TARC against two fixed-frequency PPO baselines on the
Go1: PPO-50\,Hz, the high-frequency baseline, and PPO-20\,Hz, trained at
TARC's average operating frequency. As shown in
\Cref{fig:go1_lowFreq}, TARC is equivalent to PPO-50\,Hz in unpenalized reward while operating at less
than half its frequency ($\Delta = -0.45$, TOST $p < 0.001$). Against
PPO-20\,Hz mean reward is comparable ($\Delta = +1.93$, CI $[-0.8, +5.4]$), but
TARC is far more consistent: per-seed mean reward spans $21.5$--$31.7$ for
PPO-20\,Hz against $29.7$--$30.6$ for TARC, a $12.6\times$ ratio in standard
deviation ($p = 0.008$). This
indicates that the gains arise from \emph{where} TARC spends its query
budget, not simply from querying less often. Extended results with different TARC variants are given in Appendix~\ref{sec:lowfreq}.

\paragraph{Vision-language-action model.}
\begin{figure}[ht]
    \centering
    \includegraphics[width=0.75\textwidth]{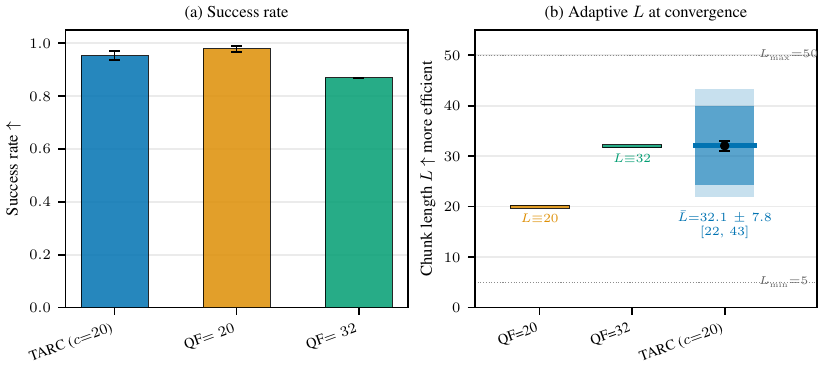}
    \caption{TARC on the LIBERO \texttt{LIVING\_ROOM\_SCENE3} environment (task 57) using $\pi_0$ as the base controller. \textbf{(a)} TARC ($c{=}20$,
    $L \in [5, 50]$) matches the success rate of the fixed
    query-frequency baseline ($\text{QF}{=}20$) and exceeds the baseline querying at $\text{QF}{=}32$, TARC's own average chunk
    length. \textbf{(b)} At convergence, TARC uses an average chunk length
    of $\bar{L} = 32.1 \pm 7.8$ (range $[22, 43]$), varying adaptively
    within episodes, compared to the fixed baseline's constant
    $L \equiv 20$.}
    \label{fig:vla}
\end{figure}
We evaluate TARC's applicability to large generative policies on the
LIBERO \texttt{LIVING\_ROOM\_SCENE3} pick-and-place environment using the
$\pi_0$ checkpoint as the base controller. Each policy query involves a
full transformer forward pass through the frozen $\pi_0$ backbone, making
the inference frequency a direct deployment cost. \Cref{fig:vla} shows
that TARC matches the success rate of the fixed chunk-size-20 $\pi_0$
baseline while using an average chunk length of $\bar{L} = 32.1 \pm 7.8$
(range $[22, 43]$) across seeds, reducing transformer forward passes by
roughly $38\%$. As a stronger comparison, we evaluate a fixed baseline
querying at $\text{QF}{=}32$, TARC's average chunk length, and find that
TARC achieves higher success rate, indicating that adaptive replanning
outperforms a fixed schedule matched to the same average budget. We show similar behavior on further tasks in Appendix~\ref{app:vla_multitask}.

%===============
%\vspace{-2pt}

%\vspace{-2pt}
\subsection{Hardware Results}
\label{sec:hardware}

We deploy our trained policies zero-shot on both hardware platforms and
evaluate task reward, control frequency, and actuation efficiency.

\paragraph{RC Car.}
\begin{figure}[ht]
    \centering
    \includegraphics[width=\textwidth]{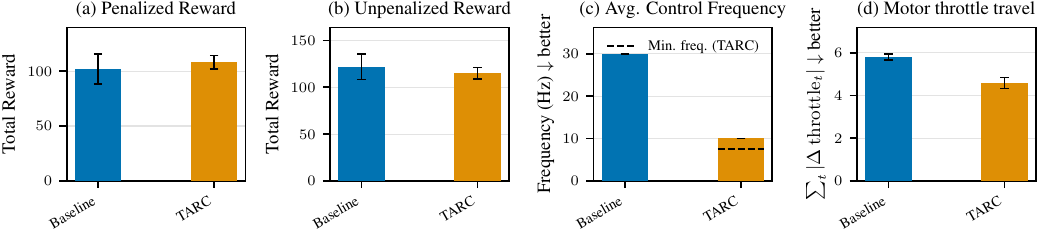}
    \caption{RC car hardware performance, mean and standard error over 5
    seeds. (a) Penalized reward ($c = 0.1$), (b) unpenalized reward,
    (c) average control frequency, (d) total motor throttle travel.}
    \label{fig:rc_car_results}
\end{figure}
With five hardware rollouts per policy we find no measurable difference in
unpenalized reward between TARC and the high-frequency baseline, while TARC exhibits
substantially smaller variance (\Cref{fig:rc_car_results}b), indicating more
consistent behavior on hardware. It further commands $21\%$ less total motor throttle
travel $\sum_t|\Delta \tau_t|$ than the fixed-frequency baseline
(\Cref{fig:rc_car_results}d). Since mechanical wear accumulates with total commanded
travel, this is a concrete deployment benefit of operating at a lower average
frequency.

\paragraph{Unitree Go1.}
\begin{figure}[ht]
    \centering
    \includegraphics[width=\textwidth]{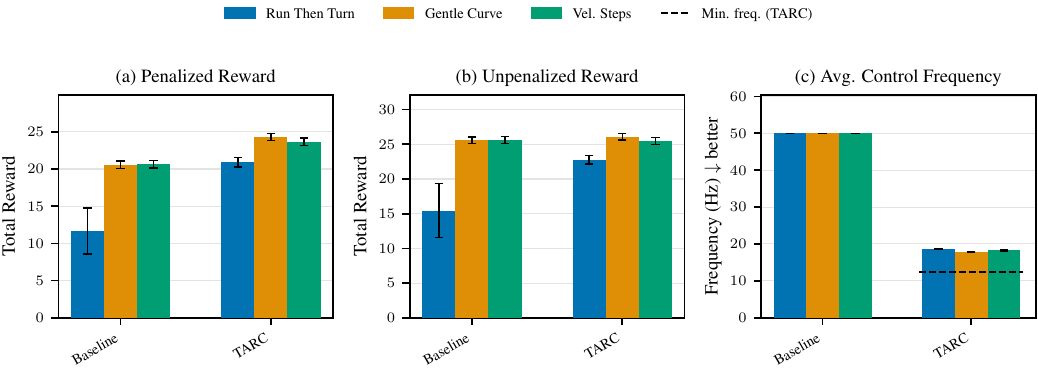}
    \caption{Go1 hardware performance across the three evaluation
    scenarios. (a) Penalized reward ($c = 0.005$), (b) unpenalized reward,
    (c) average control frequency. TARC achieves the highest penalized
    reward and the smallest variance across scenarios.}
    \label{fig:go1_results}
\end{figure}
\looseness -1 TARC exceeds the high-frequency baseline in unpenalized reward
across all three evaluation scenarios (\Cref{fig:go1_results};
$\Delta = +2.58$, CI $[+0.5, +5.2]$, $p = 0.005$), while
operating at less than half the baseline control frequency. Control
frequency responds naturally to task difficulty: lower on \textit{Gentle
Curve} and higher on \textit{Velocity Changes} and \textit{Run Then
Turn}. Because each query executes a similar ONNX graph, onboard inference cost scales linearly with query frequency. TARC issues $365 \pm 2$ policy queries per 1000 environment steps versus 1000 for PPO-50\,Hz, a $64\%$ reduction in inference load on the Go1's onboard CPU. Combined with the $21\%$ command-travel reduction on the RC car, these results demonstrate concrete deployment benefits across both platforms. Finally,
\Cref{fig:perturbations} shows TARC maintaining $16.7$\,Hz during a stable
standstill, spiking instantaneously to $50$\,Hz under an external
perturbation, then recovering immediately, state-dependent modulation
unavailable to any fixed-frequency controller.

%% file: conclusion.tex
\section{Conclusion}
\label{sec:conclusion}
%\vspace{-2pt}

We presented TARC, a reinforcement learning approach in which policies
jointly select control actions and their application durations, enabling
autonomous frequency modulation without task-specific reward redesign. We
formulate this as a constrained MDP and use a Lagrangian relaxation, 
for which \emph{any RL solver} can solve. TARC enforces user-specified query budgets reliably, and our real-world experiments on an RC car and Unitree Go1 show that it matches or exceeds high-frequency baselines
while operating at less than half their control frequency, reducing
onboard inference by $64\%$ and motor command travel by $21\%$. The
perturbation experiments reveal instantaneous state-dependent frequency
adaptation unavailable to any fixed-frequency controller. On the $\pi_0$
vision-language-action model, TARC matches the fixed query-frequency
baseline in success rate while reducing transformer forward passes
by approximately $38\%$.
%while adapting its
%replanning interval to task demand, demonstrating applicability beyond
%lightweight MLP policies.

\section{Limitations}
\label{sec:limitations}
%\vspace{-2pt}

%While the hard constraint variant eliminates manual tuning of $c$, it
%introduces the switch budget $K$ as a new hyperparameter. Although $K$
%is more interpretable, since it directly specifies the desired query fraction, transferring a budget validated in simulation to hardware remains
%untested. Validating that dual gradient ascent converges reliably under
%real-world noise is an important next step.

The time-adaptive policy uses a single global duration $\Delta t$ across all degrees of freedom. For systems with asymmetric dynamics,
such as humanoids where upper and lower body operate on different timescales,
per-subsystem duration selection would be more expressive. Additionally,
our hardware evaluations are short-horizon by design, precluding direct
measurement of energy consumption; the $64\%$ inference reduction and
$21\%$ command-travel reduction serve as concrete deployment proxies, but
translating these into power savings requires longer experiments.

Finally, TARC carries no explicit safety constraint: what counts as unsafe is only implicit in the task reward, so a reward that does not penalize unsafe states cannot induce a query when safety demands it. Since a held action is open-loop, the policy responds to a disturbance arriving mid-hold at the next decision epoch, within an interval bounded by the maximum hold duration. Coupling the query budget with explicit safety constraints, yielding a dual-constrained problem setting, is an interesting direction for future work.

%% file: appendixA.tex
\section{Proof of Proposition~\ref{prop:decomp}}
\label{app:proof}

We provide the full derivation of the per-step decomposition.

\paragraph{Setup.}
Under the time-adaptive policy class, the trajectory decomposes into decision epochs. At epoch $n$, the policy is queried at time $t_n$, outputs $(u_{t_n}, \Delta t_n)$, and the action $u_{t_n}$ is held constant for steps $t_n, t_n+1, \ldots, t_n + \Delta t_n -1$. The next query occurs at $t_{n+1} = t_n + \Delta t_n$. By definition, $q_t = 1$ if and only if $t = t_n$ for some $n$, and $q_t = 0$ otherwise. 

\paragraph{Decomposing the task reward. }
The discounted task return under the time-adaptive policy is:
\begin{align}
\mathbb{E}_\pi\!\left[\sum_t \gamma^t r(x_t, u_t)\right]
&= \mathbb{E}_\pi\!\left[\sum_n \sum_{k=0}^{\Delta t_n - 1} \gamma^{t_n + k} r(x_{t_n+k}, u_{t_n})\right] \nonumber \\
&= \mathbb{E}_\pi\!\left[\sum_n \gamma^{t_n} \sum_{k=0}^{\Delta t_n - 1} \gamma^{k} r(x_{t_n+k}, u_{t_n})\right].
\end{align}

\paragraph{Decomposing the constraint.} Since the policy is queried exactly once per epoch: 
\begin{equation}
\mathbb{E}_\pi\!\left[\sum_t \gamma^t q_t\right]
= \mathbb{E}_\pi\!\left[\sum_n \gamma^{t_n}\right].
\end{equation}

\paragraph{Combining into the Lagrangian.}
Substituting into \cref{eq:lagrangian} with $c = \lambda$:
\begin{align}
\mathcal{L}(\pi, \lambda)
&= \mathbb{E}_\pi\!\left[\sum_n \gamma^{t_n} \!\left(\sum_{k=0}^{\Delta t_n - 1} \gamma^{k} r(x_{t_n+k}, u_{t_n}) - \lambda\right)\right] + \frac{\lambda K}{1-\gamma} \nonumber \\
&= \mathbb{E}_\pi\!\left[\sum_n \gamma^{t_n} R(s_{t_n}, a_{t_n})\right] + \text{const},
\end{align}
where $R(s_{t_n}, a_{t_n}) = \bigl(\sum_{k=0}^{\Delta t_n-1} \gamma^k r(x_{t_n+k},
u_{t_n})\bigr) - c$ as in \cref{eq:reward}, and the constant $\lambda K/(1{-}\gamma)$
is independent of $\pi$. Since maximizing $\mathcal{L}$ over $\pi$ is equivalent to
maximizing $\mathbb{E}_\pi[\sum_n \gamma^{t_n} R(s_{t_n}, a_{t_n})]$, the Lagrangian
reduces to a standard MDP objective with per-decision reward $R$. \unskip\nobreak\hfill\ensuremath{\square}

\paragraph{Remark (training).} \Cref{eq:reward} is evaluated at decision epochs, so
the effective discount between consecutive epochs is $\gamma^{\Delta t_n}$. We apply
this factor in the TD residual and the GAE recursion, so the returns and advantages
optimized by PPO are the semi-Markov quantities rather than per-step ones.

%% file: appendixB.tex
\section{Experimental Details and Hyperparameters}
\label{app:hyperparams}
 
% -------------------------------------------------------
\subsection{RC Car}
\label{app:rccar}
 
\subsubsection{State Space}
 
The physical state is a 6-D vector $[p_x, p_y, \theta, v_x, v_y, \omega]$ (global position,
orientation, local velocity, angular velocity). To account for $\sim$80\,ms transmission delay,
this is augmented with the three most recent actions (12-D total). TARC policies additionally
include the episode timestep $t$ (13-D). See \Cref{tab:rccar_state_space}.
 
\begin{table}[h]
    \centering
    \caption{Components of the 13-D observation vector for the RC Car (TARC policies).
    The baseline uses the first 12 dimensions only.}
    \label{tab:rccar_state_space}
    \small
    \begin{tabular}{ll}
        \toprule
        \textbf{Dimension(s)} & \textbf{Description} \\
        \midrule
        0, 1 & Global position ($p_x, p_y$) \\
        2    & Orientation ($\theta$) \\
        3, 4 & Local velocity ($v_x, v_y$) \\
        5    & Angular velocity ($\omega$) \\
        6, 7 & Action at $t-1$ ($\delta_{t-1}, \tau_{t-1}$) \\
        8, 9 & Action at $t-2$ ($\delta_{t-2}, \tau_{t-2}$) \\
        10, 11 & Action at $t-3$ ($\delta_{t-3}, \tau_{t-3}$) \\
        12   & Episode timestep $t$ \emph{(TARC only)} \\
        \bottomrule
    \end{tabular}
\end{table}
 
\subsubsection{Action Space}
 
The action space is a 2-dimensional continuous vector $\mathbf{a} = [\delta, \tau]$,
normalized to $[-1, 1]$. The steering $\delta$ is scaled by a maximum steering angle
and the throttle $\tau$ by a maximum throttle parameter.
 
\subsubsection{Reward Function}

Following \citet{rothfuss2024bridgingsimtorealgapbayesian}, the tolerance function
$r_\text{state}$~\citep{tassa2018deepmindcontrolsuite} uses combined distance
$d_\text{total} = \sqrt{d_\text{pos}^2 + d_\theta^2}$, transformed by:
\begin{equation}
    r_\text{state}(d_\text{total}) = \frac{1}{(d_\text{total} \cdot s)^2 + 1},
\end{equation}
where $s$ is derived from the margin parameter (set to 20), giving maximum reward 1 at target and decaying smoothly with distance. 
 
\subsubsection{Simulation Dynamics}
 
The environment uses the Pacejka tire
model~\citep{Pacejka01011992, kabzan2019amzdriverlessautonomousracing} with domain
randomization~\citep{tobin2017domainrandomizationtransferringdeep} over the parameters
in \Cref{tab:rccar_dr}.
 
\begin{table}[htbp]
    \centering
    \caption{Domain randomization ranges for the RC Car simulation.}
    \label{tab:rccar_dr}
    \begin{tabular}{ll}
        \toprule
        \textbf{Parameter} & \textbf{Range} \\
        \midrule
        Mass ($m$)              & [1.6, 1.7]\,kg \\
        Inertia ($I_\text{com}$) & [0.03, 0.18] \\
        Front tire grip ($d_f$) & [0.25, 0.6] \\
        Rear tire grip ($d_r$)  & [0.15, 0.45] \\
        Motor constant ($c_{m1}$) & [10.0, 40.0] \\
        Steering limit          & [0.4, 0.75]\,rad \\
        \bottomrule
    \end{tabular}
\end{table}
 
\subsubsection{Hyperparameters}
 The following hyperparameters in \Cref{tab:rccar_hyperparams} were used for all RC Car experiments reported in \Cref{sec:setup} and \Cref{sec:hardware}. The RC Car policy is trained with PPO; TARC-specific parameters ($f_{max}$, c, and i) are set per the soft-constraint variant described in \Cref{sec:constraints}.
\begin{table}[htbp]
    \centering
    \caption{Training hyperparameters for the RC Car.}
    \label{tab:rccar_hyperparams}
    \small
    \begin{tabular}{ll}
        \toprule
        \textbf{Parameter} & \textbf{Value} \\
        \midrule
        \multicolumn{2}{l}{\textit{Environment}} \\
        Episode steps           & 200 \\
        Parallel environments   & 2048 \\
        Total timesteps         & $75 \times 10^6$ \\
        Random seeds            & 5 \\
        \midrule
        \multicolumn{2}{l}{\textit{TARC-specific}} \\
        $f_\text{max}$          & 30\,Hz \\
        Switch cost $c$         & 0.1 \\
        $f_\text{min}$ variants & $\{7.5, 6, 3\}$\,Hz \\
        \midrule
        \multicolumn{2}{l}{\textit{PPO}} \\
        Learning rate           & 3e-4 \\
        Batch size              & 1024 \\
        Minibatches             & 32 \\
        Updates per batch       & 4 \\
        Unroll length           & 10 \\
        Entropy cost            & 0.01 \\
        Discount $\gamma$       & 0.9 \\
        GAE $\lambda$           & 0.95 \\
        Clip $\epsilon$         & 0.3 \\
        Optimizer               & Adam \\
        \midrule
        \multicolumn{2}{l}{\textit{Network architecture}} \\
        Policy hidden layers    & $5 \times 32$ \\
        Critic hidden layers    & $4 \times 128$ \\
        Activation              & Swish \\
        \bottomrule
    \end{tabular}
\end{table}

 % -------------------------------------------------------
\subsection{Unitree Go1}
\label{app:go1}

\subsubsection{Environment and Policy Architecture}

We use the JoystickFlatTerrain environment from MuJoCo Playground~\cite{zakka2025mujocoplayground}, with
domain randomization over sensor noise, dynamic properties (joint calibration offsets, reflected inertia), friction and ground properties. The high-level RL policy outputs 12-D joint targets, tracked by a low-level PD controller at the underlying simulator rate.
The PD controller and reward function are identical across TARC and all baselines. TARC augments observations with the timestep $t$. Training uses an asymmetric actor-critic~\citep{pinto2017asymmetricactorcriticimagebased} setup with privileged critic state. Hardware deployment uses ONNX runtime~\citep{ONNX} on the Go1's onboard CPU. The total reward is calculated as the weighted sum of individual reward terms: $r_{total}= \sum_i w_ir_i$. We refer to MuJoCo Playground~\citep{zakka2025mujocoplayground} for reward term details.

\subsubsection{Go1 Evaluation Scenarios}
\label{app:go1_scenarios}
 
\begin{figure}[htp]
    \centering
    \includegraphics[width=\textwidth]{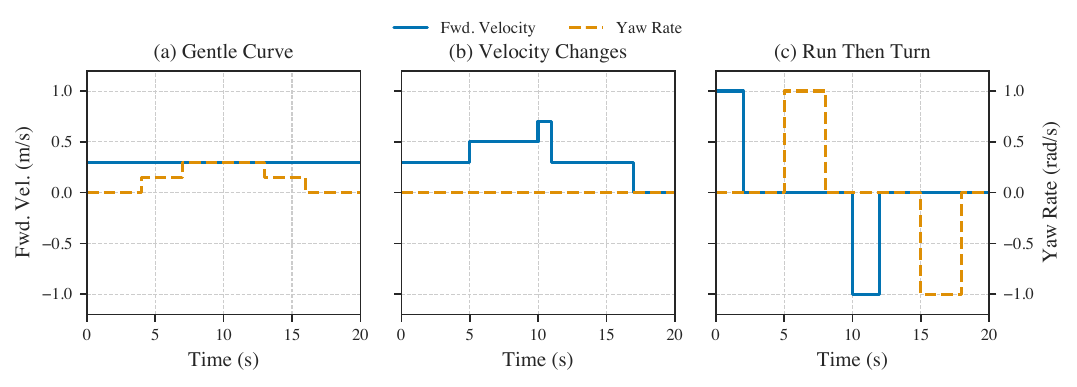}
    \caption{Go1 evaluation scenarios. \textbf{Left:} Gentle Curve. \textbf{Center:}
    Velocity Changes. \textbf{Right:} Run Then Turn.}
    \label{fig:go1_scenarios}
\end{figure}

We evaluate on three 20-second scenarios not seen during training, illustrated in \Cref{fig:go1_scenarios}:
\begin{itemize}[topsep=2pt,itemsep=1pt]
    \item \textbf{Gentle Curve:} Smooth low-speed turning; expected to favor low frequency as dynamics are slow and predictable.
    \item \textbf{Velocity Changes:} Straight walking at randomly varying commanded speeds; intermittent demand for speed corrections.
    \item \textbf{Run Then Turn:} High-speed running followed by an abrupt switch to in-place turning; tests peak reactivity requirements.
\end{itemize}
These scenarios span a range of demand intensities, from steady-state locomotion to dynamic transitions, to probe whether TARC's frequency adaptation is meaningful rather than merely averaging to a fixed rate.
 
\subsubsection{Hyperparameters}
The following hyperparameters in \Cref{tab:go1_hyperparams} were used for all Unitree Go1 experiments reported in \Cref{sec:setup} and \Cref{sec:hardware}. Default PPO settings follow the MuJoCo Playground JoystickFlatTerrain configuration; TARC-specific parameters are set per the soft-constraint variant.
\begin{table}[H]
    \centering
    \caption{Training hyperparameters for the Unitree Go1.}
    \label{tab:go1_hyperparams}
    \begin{tabular}{ll}
        \toprule
        \textbf{Parameter} & \textbf{Value} \\
        \midrule
        \multicolumn{2}{l}{\textit{Environment}} \\
        Episode steps           & 1000 \\
        Parallel environments   & 8192 \\
        Total timesteps & $600 \times 10^6$ \\
        Random seeds            & 5 \\
        \midrule
        \multicolumn{2}{l}{\textit{TARC-specific}} \\
        $f_\text{max}$          & 50\,Hz \\
        Switch cost $c$ (soft constraint)        & 0.005 \\
        $f_\text{min}$ variants & $\{12.5, 10, 5\}$\,Hz \\
        \midrule
        \multicolumn{2}{l}{\textit{PPO}} \\
        Learning rate           & 3e-4 \\
        Batch size              & 256 \\
        Minibatches             & 32 \\
        Updates per batch       & 4 \\
        Unroll length           & 20 \\
        Entropy cost            & 0.01 \\
        Discount $\gamma$       & 0.97 \\
        GAE $\lambda$           & 0.95 \\
        Clip $\epsilon$         & 0.2 \\
        Optimizer               & Adam \\
        \midrule
        \multicolumn{2}{l}{\textit{Network architecture}} \\
        Policy hidden layers    & $[512, 256, 128]$ \\
        Critic hidden layers    & $[512, 256, 128]$ \\
        Activation              & Swish \\
        \bottomrule
    \end{tabular}
    
\end{table}

%% file: appendixC.tex
\section{Supplementary Results}
\label{app:supplementary}

In this section, we present additional ablations over the hyperparameter $i$
introduced in \Cref{def:timeAdaptive}. Recall that $i$ controls the maximum
action repetition count, bounding the minimum achievable control frequency
at $f_\text{max}/i$. We evaluate three values
$i \in \{4, 5, 10\}$, corresponding to minimum frequencies of
$\{12.5, 10, 5\}$\,Hz on the Go1 and $\{7.5, 6, 3\}$\,Hz on the RC car,
and refer to each resulting controller as \textbf{TARC-$i$}. All other
hyperparameters are fixed as reported in \Cref{app:hyperparams}.

\subsection{Simulation Results}
% ---------------------------------------------------------------------
\subsubsection*{Switch Cost Sensitivity}
\label{app:switchcost}

\begin{figure}[tbp]
    \centering
    \includegraphics[width=\textwidth]{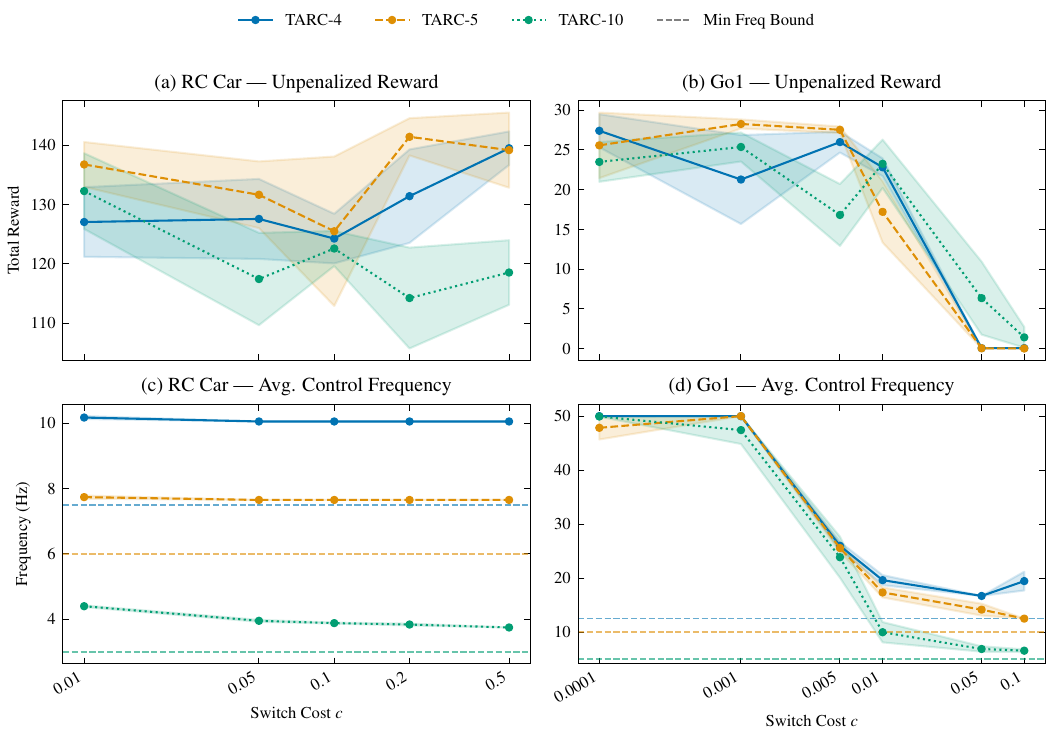}
    \caption{Switch cost sensitivity on both platforms. Left: RC car;
    right: Go1. Each pair shows unpenalized reward and average control
    frequency as a function of $c$. Dashed lines indicate the theoretical
    lower frequency bound $f_\text{max}/i$ for each TARC-$i$ variant.}
    \label{fig:switchcost}
\end{figure}

TARC introduces two platform-level hyperparameters: the maximum repetition
count $i$ (ablated throughout this section) and the per-query cost $c$
(or equivalently the Lagrange multiplier $\lambda$, see \Cref{sec:constraints}). Under the soft-constraint variant, $c$ is set once before training and directly controls the trade-off between task reward and query frequency: larger $c$ penalizes queries more heavily, yielding lower average control frequency at the potential cost of task performance. Crucially, $c$ does \emph{not} require reward redesign; it is added as a scalar to the existing per-decision reward $R$ in \Cref{prop:decomp}. \Cref{fig:switchcost} sweeps $c$ on both platforms to characterize
this sensitivity.

\paragraph{RC Car.}
The control frequency is essentially \emph{insensitive} to $c$: each TARC variant converges to a fixed operating rate (TARC-4 to 10\,Hz, TARC-5
to 7.5\,Hz, TARC-10 to 4\,Hz) across the entire range, while reward
remains high and stable. The drift-and-park task is solvable at low frequency, and the policy discovers this from dynamics alone. The switch cost acts as a directional signal rather than a precise knob.

\paragraph{Go1.}
The quadruped exhibits a clear phase transition: at $c \leq 0.001$, TARC provides no frequency benefit; between $c{=}0.001$ and $c{=}0.005$, frequency drops sharply while reward is maintained; at $c \geq 0.05$, reward collapses. At our chosen $c{=}0.005$, all variants operate well above their theoretical lower bound, confirming the agent \emph{chooses} an intermediate rate. We recommend sweeping $c$ over one decade and selecting the largest value that does not degrade unpenalized reward.

% ---------------------------------------------------------------------
\subsubsection*{Adaptive vs.\ Fixed Low-Frequency Control}
\label{sec:lowfreq}
\begin{figure}[ht]
    \centering
    \includegraphics[width=\textwidth]{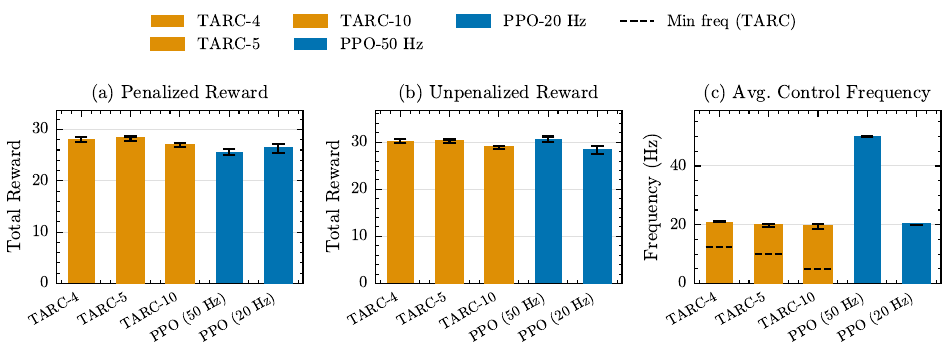}
    \caption{Adaptive (TARC) vs.\ matched fixed-frequency (PPO) control on the
    Go1. (a) Penalized reward ($c = 0.005$), (b) unpenalized
    reward, (c) average control frequency. Dashed lines in (c) indicate the
    theoretical minimum frequency of each TARC variant.}
    \label{fig:go1_lowFreq_full}
\end{figure}

\paragraph{Go1.}
\Cref{fig:go1_lowFreq} in the main paper shows results for a single TARC
variant; here we extend that comparison to all three variants
($i \in \{4, 5, 10\}$) in \Cref{fig:go1_lowFreq_full}. Across all
variants, TARC outperforms the PPO-20\,Hz baseline in mean penalized and
unpenalized reward while operating at the same average frequency, confirming
that the gain comes from adaptive allocation rather than simply running at a
lower rate. All TARC variants additionally match PPO-50\,Hz in unpenalized
reward at substantially lower control frequency, with the frequency gap
widening for larger $i$.

\begin{figure}[ht]
    \centering
    \includegraphics[width=\textwidth]{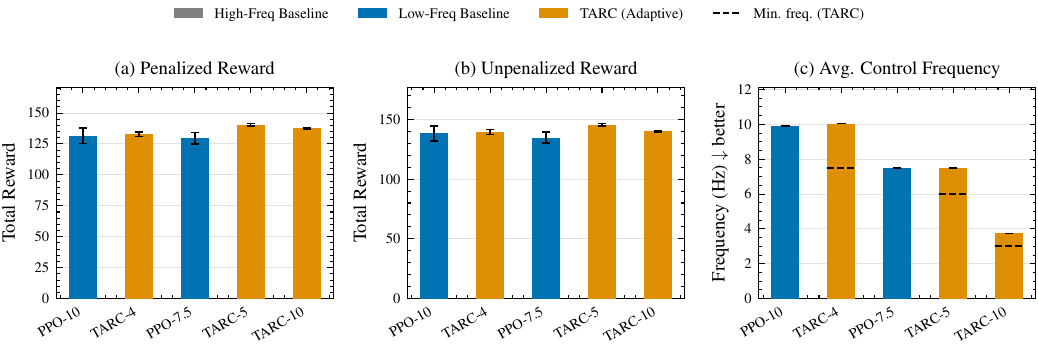}
    \caption{Adaptive (TARC) vs.\ matched fixed-frequency (PPO) control on the
    RC car drifting task. (a) Penalized reward ($c = 0.1$), (b) unpenalized
    reward, (c) average control frequency. Dashed lines in (c) indicate the
    theoretical minimum frequency of each TARC variant.}
    \label{fig:rccar_sim_comparison}
\end{figure}

\paragraph{RC Car.}
Across all matched pairs (TARC-4 vs.\ PPO-10\,Hz, TARC-5 vs.\ PPO-7.5\,Hz), TARC and the corresponding fixed-frequency baseline achieve comparable reward (\Cref{fig:rccar_sim_comparison}). The RC car drifting task exhibits a natural optimal control frequency to which both adaptive and fixed-frequency methods converge, a property of the task, not a limitation of the method.

\subsubsection*{Multi-Task VLA Generalization}
\label{app:vla_multitask}

The VLA paragraph in \Cref{sec:sim} evaluates TARC on LIBERO task 57 and demonstrates adaptive chunking behavior. Here we evaluate on two additional tasks from the \texttt{libero\_90} suite to assess whether the learned frequency adaptation generalizes across task instances.

\begin{figure}[ht]
    \centering
    \includegraphics[width=0.8\textwidth]{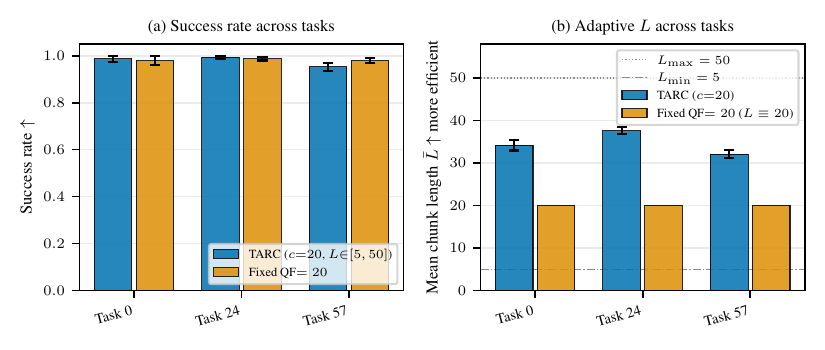}
    \caption{TARC generalization across three \texttt{libero\_90} tasks.
    \textbf{(a)} Success rate: TARC matches the fixed QF=20 baseline across all tasks.
    \textbf{(b)} Mean chunk length $\bar{L}$: TARC selects different average chunk
    lengths per task, reflecting task-specific replanning demand rather than a fixed rate.}
    \label{fig:vla_multitask}
\end{figure}

\Cref{fig:vla_multitask} shows success rate and mean chunk length $\bar{L}$ for all three tasks. Across all tasks, TARC matches the fixed QF=20 baseline in success rate ($\approx 0.96-1.0$) while operating at a higher average chunk length, reducing the number of transformer forward passes. Importantly, $\bar{L}$ varies across tasks: TARC selects longer chunks on tasks where sustained open-loop execution is sufficient, and shorter chunks where the task demands more frequent replanning. This per-task variation confirms that TARC's frequency adaptation reflects genuine task-specific demand rather than convergence to a single fixed rate.

% ---------------------------------------------------------------------
\subsection{Hardware Results}

\subsubsection*{Go1 Hardware Results} 
\begin{figure}[htp]
    \centering
    \includegraphics[width=0.9\textwidth]{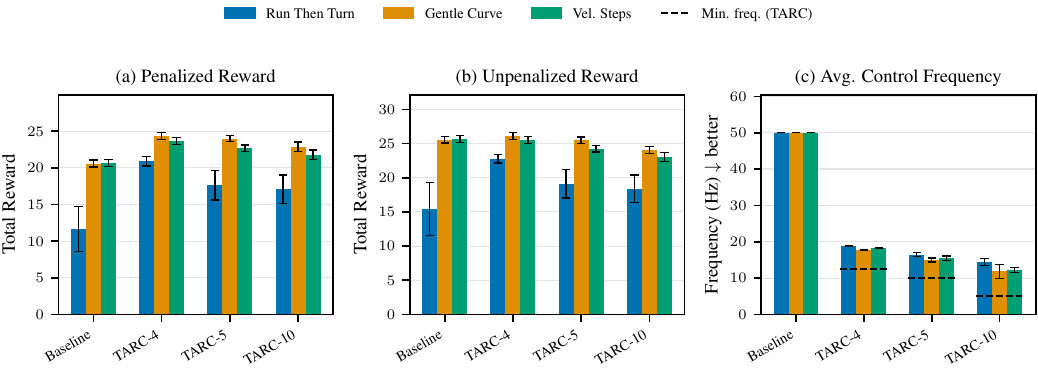}
    \caption{Hardware results ablation for different $i$. The different TARC variants perform consistently on the different tasks compared to the baseline. Mean and standard error over 5 seeds.}
    \label{fig:go1_hw_full}
\end{figure}

We evaluate all three TARC variants ($i \in \{4, 5, 10\}$) zero-shot on
the hardware scenarios introduced in \Cref{app:go1_scenarios}; results are
in \Cref{fig:go1_hw_full}. All variants perform consistently across
scenarios and match or exceed the high-frequency baseline in unpenalized
reward, confirming that the simulation results in \Cref{sec:hardware}
are not specific to the single variant reported in the main paper.

\subsubsection*{Sim-to-Real Transfer on the RC Car} 
\begin{figure}[H]
    \centering
    \includegraphics[width=0.9\textwidth]{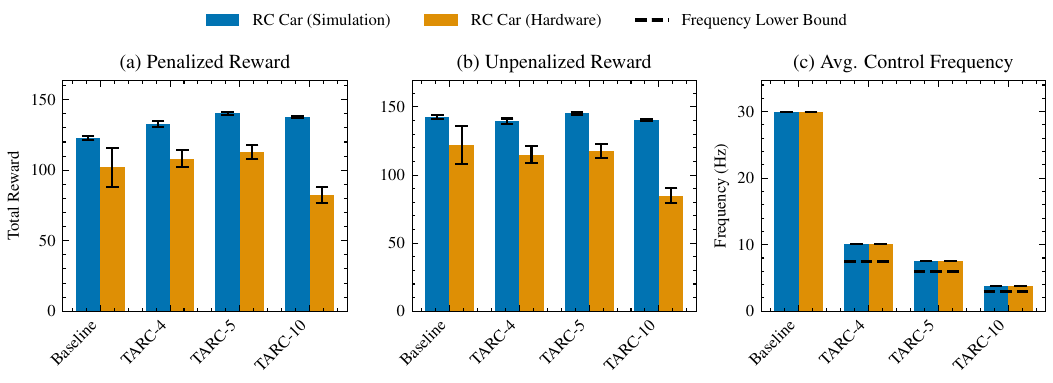}
    \caption{Simulation vs.\ hardware comparison for each TARC variant on the RC
    car. (a) Penalized reward, (b) unpenalized reward, (c) average control
    frequency. Frequencies transfer almost perfectly; reward gaps grow only for
    TARC-10, consistent with open-loop sensitivity at long repetition horizons.
    Mean and standard error over 5 seeds.}
    \label{fig:rc_car_simhw}
\end{figure}

\Cref{fig:rc_car_simhw} compares simulation and hardware performance for
each TARC variant on the RC car. Two findings stand out. 
(1)~\textbf{Frequency transfer} — average frequencies selected in
simulation are nearly identical to those on hardware, confirming that the
learned query schedule is not an artifact of simulator dynamics and
transfers without modification.
(2)~\textbf{Reward gap vs.\ repetition horizon} — the sim-to-real reward
gap is negligible for TARC-4 and TARC-5 but grows substantially for
TARC-10. At $i{=}10$, a single action is held for up to $\approx$330\,ms,
during which unmodeled disturbances compound without correction. This
suggests $i \leq 5$ as a practical upper bound for sim-to-real transfer
on fast dynamic platforms.

% ---------------------------------------------------------------------

%% file: example.bib
@misc{wang2024reinforcementlearningelastictime,
      title={Reinforcement Learning with Elastic Time Steps}, 
      author={Dong Wang and Giovanni Beltrame},
      year={2024},
      eprint={2402.14961},
      archivePrefix={arXiv},
      primaryClass={cs.RO},
      url={https://arxiv.org/abs/2402.14961}, 
}

@misc{wang2024moseacstreamlinedvariabletime,
      title={MOSEAC: Streamlined Variable Time Step Reinforcement Learning}, 
      author={Dong Wang and Giovanni Beltrame},
      year={2024},
      eprint={2406.01521},
      archivePrefix={arXiv},
      primaryClass={cs.LG},
      url={https://arxiv.org/abs/2406.01521}, 
}

@misc{li2025gaitnetaugmentedimplicitkinodynamicmpc,
      title={Gait-Net-augmented Implicit Kino-dynamic MPC for Dynamic Variable-frequency Humanoid Locomotion over Discrete Terrains}, 
      author={Junheng Li and Ziwei Duan and Junchao Ma and Quan Nguyen},
      year={2025},
      eprint={2502.02934},
      archivePrefix={arXiv},
      primaryClass={cs.RO},
      url={https://arxiv.org/abs/2502.02934}, 
}

@ARTICLE{10806807,
  author={Nguyen, Chuong and Altawaitan, Abdullah and Duong, Thai and Atanasov, Nikolay and Nguyen, Quan},
  journal={IEEE Robotics and Automation Letters}, 
  title={Variable-Frequency Model Learning and Predictive Control for Jumping Maneuvers on Legged Robots}, 
  year={2025},
  volume={10},
  number={2},
  pages={1321-1328},
  doi={10.1109/LRA.2024.3519864}}

@inproceedings{park2022timediscretizationinvariantsafeaction,
 author = {Park, Seohong and Kim, Jaekyeom and Kim, Gunhee},
 booktitle = {Advances in Neural Information Processing Systems},
 editor = {M. Ranzato and A. Beygelzimer and Y. Dauphin and P.S. Liang and J. Wortman Vaughan},
 pages = {267--279},
 publisher = {Curran Associates, Inc.},
 title = {Time Discretization-Invariant Safe Action Repetition for Policy Gradient Methods},
 url = {https://proceedings.neurips.cc/paper_files/paper/2021/file/024677efb8e4aee2eaeef17b54695bbe-Paper.pdf},
 volume = {34},
 year = {2021}
}

@INPROCEEDINGS{10342408,
  author={Karimi, Amirmohammad and Jin, Jun and Luo, Jun and Mahmood, A. Rupam and Jagersand, Martin and Tosatto, Samuele},
  booktitle={2023 IEEE/RSJ International Conference on Intelligent Robots and Systems (IROS)}, 
  title={Dynamic Decision Frequency with Continuous Options}, 
  year={2023},
  volume={},
  number={},
  pages={7545-7552},
  doi={10.1109/IROS55552.2023.10342408}}

@InProceedings{pmlr-v139-amin21a,
  title = 	 {Locally Persistent Exploration in Continuous Control Tasks with Sparse Rewards},
  author =       {Amin, Susan and Gomrokchi, Maziar and Aboutalebi, Hossein and Satija, Harsh and Precup, Doina},
  booktitle = 	 {Proceedings of the 38th International Conference on Machine Learning},
  pages = 	 {275--285},
  year = 	 {2021},
  editor = 	 {Meila, Marina and Zhang, Tong},
  volume = 	 {139},
  series = 	 {Proceedings of Machine Learning Research},
  month = 	 {18--24 Jul},
  publisher =    {PMLR},
  url = 	 {https://proceedings.mlr.press/v139/amin21a.html},
}

@INPROCEEDINGS{10160357,
  author={Gangapurwala, Siddhant and Campanaro, Luigi and Havoutis, Ioannis},
  booktitle={2023 IEEE International Conference on Robotics and Automation (ICRA)}, 
  title={Learning Low-Frequency Motion Control for Robust and Dynamic Robot Locomotion}, 
  year={2023},
  volume={},
  number={},
  pages={5085-5091},
  doi={10.1109/ICRA48891.2023.10160357}}

@inproceedings{sharma2017learning,
  title={Learning to Repeat: Fine Grained Action Repetition for Deep Reinforcement Learning},
  author={Sharma, Sahil and Lakshminarayanan, Aravind S and Ravindran, Balaraman},
  booktitle={International Conference on Learning Representations},
  year={2017}
}

@InProceedings{pmlr-v119-metelli20a,
  title = 	 {Control Frequency Adaptation via Action Persistence in Batch Reinforcement Learning},
  author =       {Metelli, Alberto Maria and Mazzolini, Flavio and Bisi, Lorenzo and Sabbioni, Luca and Restelli, Marcello},
  booktitle = 	 {Proceedings of the 37th International Conference on Machine Learning},
  pages = 	 {6862--6873},
  year = 	 {2020},
  editor = 	 {III, Hal Daumé and Singh, Aarti},
  volume = 	 {119},
  series = 	 {Proceedings of Machine Learning Research},
  month = 	 {13--18 Jul},
  publisher =    {PMLR},
  url = 	 {https://proceedings.mlr.press/v119/metelli20a.html},
}

@inproceedings{
    Hafner2020Dream,
    title={Dream to Control: Learning Behaviors by Latent Imagination},
    author={Danijar Hafner and Timothy Lillicrap and Jimmy Ba and Mohammad Norouzi},
    booktitle={International Conference on Learning Representations},
    year={2020},
    url={https://openreview.net/forum?id=S1lOTC4tDS}
}

@inproceedings{braylan:aaai15ws,
title={Frame Skip Is a Powerful Parameter for Learning to Play Atari},
author={Alexander Braylan and Mark Hollenbeck and Elliot Meyerson and Risto Miikkulainen},
booktitle={AAAI-15 Workshop on Learning for General Competency in Video Games},
url="http://nn.cs.utexas.edu/?braylan:aaai15ws",
year={2015}
}

@INPROCEEDINGS{6425820,
  author={Heemels, W.P.M.H. and Johansson, K.H. and Tabuada, P.},
  booktitle={2012 IEEE 51st IEEE Conference on Decision and Control (CDC)}, 
  title={An introduction to event-triggered and self-triggered control}, 
  year={2012},
  volume={},
  number={},
  pages={3270-3285},
  doi={10.1109/CDC.2012.6425820}}

@ARTICLE{5411835,
  author={Anta, Adolfo and Tabuada, Paulo},
  journal={IEEE Transactions on Automatic Control}, 
  title={To Sample or not to Sample: Self-Triggered Control for Nonlinear Systems}, 
  year={2010},
  volume={55},
  number={9},
  pages={2030-2042},
  doi={10.1109/TAC.2010.2042980}}

@inproceedings{treven2024sensecontroltimeadaptiveapproach,
 author = {Treven, Lenart and Sukhija, Bhavya and As, Yarden and D\"{o}rfler, Florian and Krause, Andreas},
 booktitle = {Advances in Neural Information Processing Systems},
 editor = {A. Globerson and L. Mackey and D. Belgrave and A. Fan and U. Paquet and J. Tomczak and C. Zhang},
 pages = {63654--63685},
 publisher = {Curran Associates, Inc.},
 title = {When to Sense and Control? A Time-adaptive Approach for Continuous-Time RL},
 url = {https://proceedings.neurips.cc/paper_files/paper/2024/file/746b0a1e6e3cff8d968ba6d2e6fff049-Paper-Conference.pdf},
 volume = {37},
 year = {2024}
}

@inproceedings{
freeman2021braxdifferentiablephysics,
title={Brax - A Differentiable Physics Engine for Large Scale Rigid Body Simulation},
author={C. Daniel Freeman and Erik Frey and Anton Raichuk and Sertan Girgin and Igor Mordatch and Olivier Bachem},
booktitle={Thirty-fifth Conference on Neural Information Processing Systems Datasets and Benchmarks Track (Round 1)},
year={2021},
url={https://openreview.net/forum?id=VdvDlnnjzIN}
}

@misc{schulman2017proximalpolicyoptimizationalgorithms,
      title={Proximal Policy Optimization Algorithms}, 
      author={John Schulman and Filip Wolski and Prafulla Dhariwal and Alec Radford and Oleg Klimov},
      year={2017},
      eprint={1707.06347},
      archivePrefix={arXiv},
      primaryClass={cs.LG},
      url={https://arxiv.org/abs/1707.06347}, 
}

@misc{OptiTrackRobotics,
  title        = {{Robotics Motion Capture Systems}},
  author       = {{OptiTrack}},
  howpublished = {\url{https://optitrack.com/applications/robotics/}},
  year         = {2024},
  note         = {Accessed: July 22, 2024}
}

@article{kabzan2019amzdriverlessautonomousracing,
author = {Kabzan, Juraj and Valls, Miguel I. and Reijgwart, Victor J. F. and Hendrikx, Hubertus F. C. and Ehmke, Claas and Prajapat, Manish and Bühler, Andreas and Gosala, Nikhil and Gupta, Mehak and Sivanesan, Ramya and Dhall, Ankit and Chisari, Eugenio and Karnchanachari, Napat and Brits, Sonja and Dangel, Manuel and Sa, Inkyu and Dubé, Renaud and Gawel, Abel and Pfeiffer, Mark and Liniger, Alexander and Lygeros, John and Siegwart, Roland},
title = {AMZ Driverless: The full autonomous racing system},
journal = {Journal of Field Robotics},
volume = {37},
number = {7},
pages = {1267-1294},
doi = {https://doi.org/10.1002/rob.21977},
url = {https://onlinelibrary.wiley.com/doi/abs/10.1002/rob.21977},
eprint = {https://onlinelibrary.wiley.com/doi/pdf/10.1002/rob.21977},
year = {2020}
}

@article{Pacejka01011992,
author = {Hans B. Pacejka and Egbert Bakker},
title = {THE MAGIC FORMULA TYRE MODEL},
journal = {Vehicle System Dynamics},
volume = {21},
number = {sup001},
pages = {1--18},
year = {1992},
publisher = {Taylor \& Francis},
doi = {10.1080/00423119208969994},
URL = {https://doi.org/10.1080/00423119208969994},
eprint = {https://doi.org/10.1080/00423119208969994}}

@ARTICLE{bhardwaj2024dataefficienttaskgeneralizationprobabilistic,
  author={Bhardwaj, Arjun and Rothfuss, Jonas and Sukhija, Bhavya and As, Yarden and Hutter, Marco and Coros, Stelian and Krause, Andreas},
  journal={IEEE Robotics and Automation Letters}, 
  title={Data-Efficient Task Generalization via Probabilistic Model-Based Meta Reinforcement Learning}, 
  year={2024},
  volume={9},
  number={4},
  pages={3918-3925},
  doi={10.1109/LRA.2024.3371260}}

@INPROCEEDINGS{sukhija2023gradientbasedtrajectoryoptimizationlearned,
  author={Sukhija, Bhavya and Köhler, Nathanael and Zamora, Miguel and Zimmermann, Simon and Curi, Sebastian and Krause, Andreas and Coros, Stelian},
  booktitle={2023 IEEE International Conference on Robotics and Automation (ICRA)}, 
  title={Gradient-Based Trajectory Optimization With Learned Dynamics}, 
  year={2023},
  volume={},
  number={},
  pages={1011-1018},
  doi={10.1109/ICRA48891.2023.10161574}}

@INPROCEEDINGS{tobin2017domainrandomizationtransferringdeep,
  author={Tobin, Josh and Fong, Rachel and Ray, Alex and Schneider, Jonas and Zaremba, Wojciech and Abbeel, Pieter},
  booktitle={2017 IEEE/RSJ International Conference on Intelligent Robots and Systems (IROS)}, 
  title={Domain randomization for transferring deep neural networks from simulation to the real world}, 
  year={2017},
  volume={},
  number={},
  pages={23-30},
  doi={10.1109/IROS.2017.8202133}}

@INPROCEEDINGS{rothfuss2024bridgingsimtorealgapbayesian,
  author={Rothfuss, Jonas and Sukhija, Bhavya and Treven, Lenart and Dörfler, Florian and Coros, Stelian and Krause, Andreas},
  booktitle={2024 IEEE/RSJ International Conference on Intelligent Robots and Systems (IROS)}, 
  title={Bridging the Sim-to-Real Gap with Bayesian Inference}, 
  year={2024},
  volume={},
  number={},
  pages={10784-10791},
  doi={10.1109/IROS58592.2024.10801505}}

@misc{tassa2018deepmindcontrolsuite,
      title={DeepMind Control Suite}, 
      author={Yuval Tassa and Yotam Doron and Alistair Muldal and Tom Erez and Yazhe Li and Diego de Las Casas and David Budden and Abbas Abdolmaleki and Josh Merel and Andrew Lefrancq and Timothy Lillicrap and Martin Riedmiller},
      year={2018},
      eprint={1801.00690},
      archivePrefix={arXiv},
      primaryClass={cs.AI},
      url={https://arxiv.org/abs/1801.00690}, 
}

@misc{zakka2025mujocoplayground,
      title={MuJoCo Playground}, 
      author={Kevin Zakka and Baruch Tabanpour and Qiayuan Liao and Mustafa Haiderbhai and Samuel Holt and Jing Yuan Luo and Arthur Allshire and Erik Frey and Koushil Sreenath and Lueder A. Kahrs and Carmelo Sferrazza and Yuval Tassa and Pieter Abbeel},
      year={2025},
      eprint={2502.08844},
      archivePrefix={arXiv},
      primaryClass={cs.RO},
      url={https://arxiv.org/abs/2502.08844}, 
}

@misc{ONNX,
  title        = {{ONNX: Open Neural Network Exchange}},
  author       = {{ONNX Community}},
  howpublished = {\url{https://onnx.ai/}},
  year         = {2024},
  note         = {Accessed: July 22, 2024}
}

@misc{UnitreeGo1,
  title        = {{Go1 Quadruped Robot}},
  author       = {{Unitree Robotics}},
  howpublished = {\url{https://www.unitree.com/go1}},
  year         = {2024},
  note         = {Accessed: July 22, 2024}
}

@misc{pinto2017asymmetricactorcriticimagebased,
      title={Asymmetric Actor Critic for Image-Based Robot Learning}, 
      author={Lerrel Pinto and Marcin Andrychowicz and Peter Welinder and Wojciech Zaremba and Pieter Abbeel},
      year={2017},
      eprint={1710.06542},
      archivePrefix={arXiv},
      primaryClass={cs.RO},
      url={https://arxiv.org/abs/1710.06542}, 
}

@misc{zhao2023learningfinegrainedbimanualmanipulation,
      title={Learning Fine-Grained Bimanual Manipulation with Low-Cost Hardware}, 
      author={Tony Z. Zhao and Vikash Kumar and Sergey Levine and Chelsea Finn},
      year={2023},
      eprint={2304.13705},
      archivePrefix={arXiv},
      primaryClass={cs.RO},
      url={https://arxiv.org/abs/2304.13705}, 
}

@misc{chi2024diffusionpolicyvisuomotorpolicy,
      title={Diffusion Policy: Visuomotor Policy Learning via Action Diffusion}, 
      author={Cheng Chi and Zhenjia Xu and Siyuan Feng and Eric Cousineau and Yilun Du and Benjamin Burchfiel and Russ Tedrake and Shuran Song},
      year={2024},
      eprint={2303.04137},
      archivePrefix={arXiv},
      primaryClass={cs.RO},
      url={https://arxiv.org/abs/2303.04137}, 
}

@misc{black2026pi0visionlanguageactionflowmodel,
      title={$\pi_0$: A Vision-Language-Action Flow Model for General Robot Control}, 
      author={Kevin Black and Noah Brown and Danny Driess and Adnan Esmail and Michael Equi and Chelsea Finn and Niccolo Fusai and Lachy Groom and Karol Hausman and Brian Ichter and Szymon Jakubczak and Tim Jones and Liyiming Ke and Sergey Levine and Adrian Li-Bell and Mohith Mothukuri and Suraj Nair and Karl Pertsch and Lucy Xiaoyang Shi and James Tanner and Quan Vuong and Anna Walling and Haohuan Wang and Ury Zhilinsky},
      year={2026},
      eprint={2410.24164},
      archivePrefix={arXiv},
      primaryClass={cs.LG},
      url={https://arxiv.org/abs/2410.24164}, 
}

@misc{liang2026adaptiveactionchunkinginferencetime,
      title={Adaptive Action Chunking at Inference-time for Vision-Language-Action Models}, 
      author={Yuanchang Liang and Xiaobo Wang and Kai Wang and Shuo Wang and Xiaojiang Peng and Haoyu Chen and David Kim Huat Chua and Prahlad Vadakkepat},
      year={2026},
      eprint={2604.04161},
      archivePrefix={arXiv},
      primaryClass={cs.RO},
      url={https://arxiv.org/abs/2604.04161}, 
}

@misc{liu2023liberobenchmarkingknowledgetransfer,
      title={LIBERO: Benchmarking Knowledge Transfer for Lifelong Robot Learning}, 
      author={Bo Liu and Yifeng Zhu and Chongkai Gao and Yihao Feng and Qiang Liu and Yuke Zhu and Peter Stone},
      year={2023},
      eprint={2306.03310},
      archivePrefix={arXiv},
      primaryClass={cs.AI},
      url={https://arxiv.org/abs/2306.03310}, 
}

@misc{wagenmaker2025steeringdiffusionpolicylatent,
      title={Steering Your Diffusion Policy with Latent Space Reinforcement Learning}, 
      author={Andrew Wagenmaker and Mitsuhiko Nakamoto and Yunchu Zhang and Seohong Park and Waleed Yagoub and Anusha Nagabandi and Abhishek Gupta and Sergey Levine},
      year={2025},
      eprint={2506.15799},
      archivePrefix={arXiv},
      primaryClass={cs.RO},
      url={https://arxiv.org/abs/2506.15799}, 
}
